\documentclass[pdflatex,sn-mathphys-num]{sn-jnl}%

\usepackage{graphicx}%
\usepackage{multirow}%
\usepackage{amsmath,amssymb,amsfonts}%
\usepackage{amsthm}%
\usepackage{mathrsfs}%
\usepackage[title]{appendix}%
\usepackage{xcolor}%
\usepackage{textcomp}%
\usepackage{manyfoot}%
\usepackage{booktabs}%
\usepackage{algorithm}%
\usepackage{algorithmicx}%
\usepackage{algpseudocode}%
\usepackage{listings}%

\begin{document}

\title[	CrystalX: High-accuracy Crystal Structure Analysis Using Deep Learning]{	CrystalX: High-accuracy Crystal Structure Analysis Using Deep Learning}

\author[1,2,3]{\fnm{Kaipeng} \sur{Zheng}}

\author*[1,2,3]{\fnm{Weiran} \sur{Huang}}\email{weiran.huang@sjtu.edu.cn}

\author[2]{\fnm{Wanli} \sur{Ouyang}}

\author*[2,3]{\fnm{Han-Sen} \sur{Zhong}}\email{zhonghansen@pjlab.org.cn}

\author*[2,3]{\fnm{Yuqiang} \sur{Li}}\email{liyuqiang@pjlab.org.cn}

\affil[1]{\orgdiv{School of Computer Science}, \orgname{Shanghai Jiao Tong University}, \orgaddress{\city{Shanghai}, \postcode{200240}, \country{China}}}

\affil[2]{\orgname{Shanghai Artificial Intelligence Laboratory}, \orgaddress{\city{Shanghai}, \postcode{200232}, \country{China}}}

\affil[3]{\orgname{Shanghai Innovation Institute}, \orgaddress{\city{Shanghai}, \postcode{200230}, \country{China}}}

\abstract{

Atomic structure analysis of crystalline materials is a paramount endeavor in both chemical and material sciences. 
This sophisticated technique necessitates not only a solid foundation in crystallography but also a profound comprehension of the intricacies of the accompanying software, posing a significant challenge in meeting the rigorous daily demands. 
For the first time, we confront this challenge head-on by harnessing the power of deep learning for fully automated routine structure analysis at the full-atom level.
To validate the performance of the model, named CrystalX, we employed a dataset comprising over 50,000 X-ray diffraction measurements derived from authentic experiments. Under a strict temporal validation scheme that separates training and test data by publication time, CrystalX substantially outperformed the automated baseline and adept at deciphering intricate geometric patterns. 
Remarkably, CrystalX revealed that even peer-reviewed publications harbor expert interpretation errors that can evade stringent CheckCIF A/B-level alerts, yet CrystalX adeptly rectifies them.
It has already been successfully applied in our day-to-day pipeline, enabling fully automated, human-free structure analysis for newly discovered compounds.
Overall, CrystalX marks the beginning of a new era in automating routine structural analysis within self-driving laboratories.
}

\maketitle

\section*{Introduction}

Crystal structure analysis, as a pivotal technical tool to gain valuable insights into the microscopic realm of matter, has significantly transformed humanity's perception of the world, helping scientists ascertain the structures of vital substances, such as nucleic acids~\cite{dna}, penicillin~\cite{penicillin}, vitamin B12~\cite{vitamin}, and insulin~\cite{insulin}. These breakthroughs have fueled major scientific advancements. 
As technology progresses with automated synthesis~\cite{li2015synthesis,auto-synthesis1,auto-synthesis2,blair2022automated-synthesis,wang2024rapid-synthesis} and high-throughput crystallization~\cite{auto-crystal1,auto-crystal2,encapsu-crystal}, the demand for routine structure analysis has surged. 
Yet, this task remains both labor-intensive and time-consuming, relying on the specialized skills of crystallographers.

Establishing a routine method to automate daily structure analysis with both precision and speed has long been a key objective to accelerate chemical discoveries.
Routine structure analysis generally involves phasing and interpreting coarse electron density.
Phasing diffraction data into coarse electron density, has become highly automated, thanks to continuous advancements in modern crystallographic software~\cite{shelxt,shelxs,charge-flipping,altomare1999sir97,sir2014}. 
Recent breakthroughs in deep learning-based phasing~\cite{phai} have further improved this.
However, the automation of routine structure analysis continues to be impeded by the second step: the accurate interpretation of coarse electron density.
Converting an initial density map into a chemically and crystallographically consistent atomic model often requires iterative decisions, including assigning element types, placing hydrogen atoms, and refining the model based on residual density and validation feedback. As a result, routine analysis in many settings still relies substantially on expert judgment and interactive use of specialized refinement and visualization software~\cite{shelxl,altomare1999sir97,betteridge2003crystals,olex2,hubschle2011shelxle,farrugia1999wingx}.

\begin{figure}[!tbp]
    \centering
    \includegraphics[width=\textwidth, trim={0cm 0cm 0cm 0}, clip]{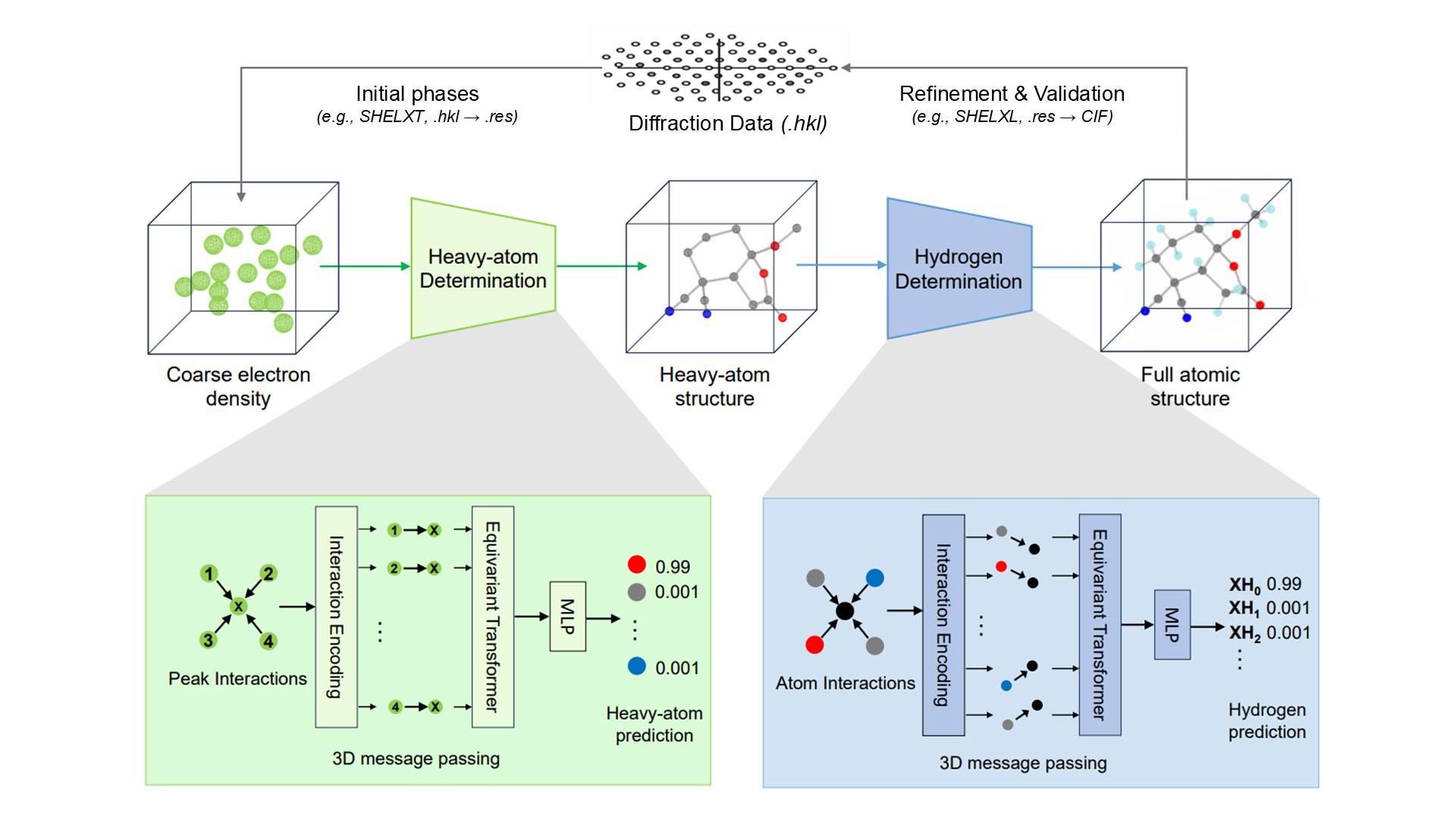}
    \caption{\textbf{The CrystalX neural network approach to crystal structure analysis.}
    Diffraction measurements are automatically phased (e.g., SHELXT) to produce coarse electron-density peak maps, which serve as input to CrystalX. An Equivariant Transformer first models 3D peak–peak interactions to identify heavy (non-hydrogen) atom positions and yield a heavy-atom framework; a second Equivariant Transformer then models atom–atom interactions on this framework to predict hydrogen locations, producing the full atomic structure, which can subsequently be refined with SHELXL.
    }
    \label{fig:fig1}
\end{figure}

We show that artificial intelligence (AI) holds the potential to overcome this challenge.
Through deep learning, AI has become a formidable tool for automating routine tasks and advancing scientific discovery \cite{ai4med,ai4mat,ai4smell,alphafold1,tubiana2022scannet}. 
Particularly, pioneering studies in areas such as computer-aided diagnosis \cite{ai4med}, material innovation \cite{ai4mat}, and molecular property prediction \cite{ai4smell} have underscored AI's capability in extracting hidden information from complex patterns that are challenging to discern manually.
In crystallography, the potential of AI for interpreting electron density was first highlighted in the late 1970s~\cite{early-ml}.
More recently, deep learning has been applied to analyze crystal symmetry in powder diffraction~\cite{park2017classification,ziletti2018insightful,oviedo2019fast} and electron diffraction~\cite{kaufmann2020crystal}. 
Furthermore, recent research also highlights its promise in phasing low-quality X-ray diffraction data~\cite{phai}. 
Nonetheless, the application of AI to fully automate routine structure analysis remains unexplored.

This work introduces the first deep learning model, CrystalX, to fully automate routine structure analysis with high accuracy, bringing the time required for the entire process down to seconds.
We validate CrystalX on 51,334 real experimental XRD patterns from the Crystallography Open Database (COD)~\cite{cod}, spanning organic, organometallic, and inorganic compounds across 83 elements and 86 space groups. 
Under the strict time-ordered test split, CrystalX achieves 99.71\% accuracy for non-hydrogen atoms and 99.42\% for hydrogen atoms, and reaches high structure-level integrity (all atoms correct). CrystalX further delivers calibrated uncertainty, enabling lightweight probability-guided corrections that improve structure integrity without expensive search. On challenging benchmarks targeting the regimes where automation most often fails---low-SNR measurements and large, compositionally diverse structures---CrystalX substantially outperforms the widely used Olex2~\cite{olex2} automated workflow, and its solutions closely match expert-solved structures under CheckCIF~\cite{platon} validation. CrystalX scales to exceptionally large structures (up to 370 non-hydrogen atoms) within seconds, supporting high-throughput crystallization-to-structure pipelines.
Additionally, we investigate CrystalX's intrinsic behavior, confirming the high interpretability of the learned geometric relationships.
CrystalX has already demonstrated its practical application. For example, 9 errors have been identified and corrected from 1,559 instances in Journal Citation Reports (JCR)~\cite{jcr} Q1 journals, revealing that even peer-reviewed journals contain errors that escape both human detection and the stringent CheckCIF A/B alert criteria.
Finally, CrystalX has already been integrated into our day-to-day crystallography workflow for newly discovered compounds. 
We directly compared it with the latest version of the automated structure-solution system, AutoChem (ac7)~\cite{diffraction2019crysalispro}. 
Database-scale benchmarking was not feasible, as AutoChem depends on instrument-generated files that are not available in published crystallographic repositories, limiting evaluation to prospective real-world experiments. 
In head-to-head tests on two newly discovered compounds, together with an additional recently published case identified through a broad literature search~\cite{autochem_case}, CrystalX achieved fully automated and correct structure analysis in all three cases while also running faster. By contrast, regardless of the configuration, AutoChem was able to solve at most one of the three compounds correctly.

\section*{Results}
\subsection*{The neural network approach}
Crystal structure analysis typically begins by phasing experimental X-ray diffraction data into an initial, coarse electron density map. 
Our approach is designed to predict the complete atomic structure from the coarse electron density that is automatically produced by routine software~\cite{shelxt,shelxs,charge-flipping,altomare1999sir97}.
To achieve this objective, we split it into a pipeline that separately determines the heavy atoms (i.e., non-hydrogen atoms) and the hydrogen atoms, as depicted in Fig.~\ref{fig:fig1}.
The coarse electron density can be represented as a point cloud, where each peak corresponds to a specific location with a charge density value.
Distinguishing elements with similar atomic numbers, such as the ubiquitous boron, carbon, nitrogen, oxygen, and fluorine, based on charge density can be challenging. 
This difficulty is particularly pronounced in real-world experimental measurements where data quality may not be optimal.
The foundation for elemental determination hinges on precisely capturing atomic geometric interaction patterns, such as distances, angles, and dihedral angles, derived from electron density peaks. 
With the advancements in geometric deep learning~\cite{schutt2017schnet,dimenet,spherenet,wang2022comenet,torchmd}, these complex patterns can now be learned by deep-learning models in an end-to-end manner, removing the need for rule-based design or intricate handcrafted features.
We harness the power of an advanced Equivariant Transformer model, TorchMD-NET~\cite{torchmd}, to decode geometric interaction patterns from electron density peaks, enabling accurate identification of heavy elements.
We also show that various geometric deep learning methods can effectively address this problem, with TorchMD-NET emerging as a particularly competitive option due to its superior balance of accuracy and efficiency, making it a highly practical solution.

\begin{figure}[!t]
    \centering
    \includegraphics[width=\textwidth, trim={0cm 2cm 0cm 0}, clip]{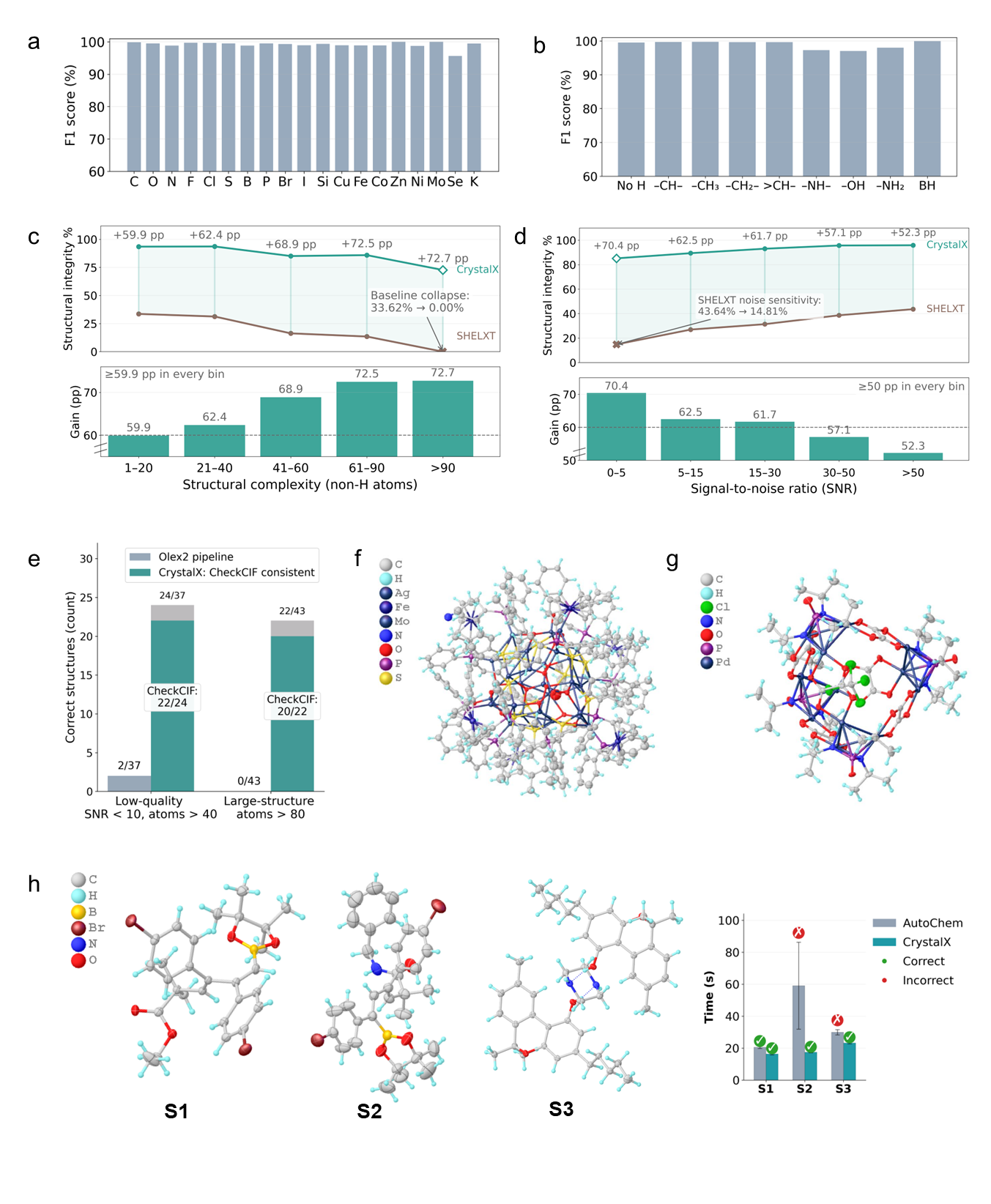}
    \caption{\textbf{Performance overview on large-scale experimental XRD data and newly discovered compounds.} \textbf{(a)} Element-level F1 scores for common non-hydrogen elements. \textbf{(b)} F1 scores for common hydrogen assignments. \textbf{(c, d)} Structure-integrity accuracy (all atoms correct) for non-hydrogen identification as a function of (c) structural complexity and (d) diffraction quality (SNR). \textbf{(e)} End-to-end all-atom structure determination on low-quality and large-structure subsets, with CheckCIF validation against expert-solved CIFs. \textbf{(f, g)} Automated analysis of exceptionally large, complex structures (up to 370 non-hydrogen atoms). \textbf{(h)} Fully automated structure analysis of newly discovered compounds in our routine workflow, as well as a recently published structure, benchmarked against AutoChem.}
    \label{fig:fig2}
\end{figure}

Hydrogen atoms are often indistinct in electron density maps obtained from traditional X-ray diffraction~\cite{2017hydrogen,2019hydrogen}. 
The critical aspect of determining hydrogen atoms involves accurately identifying the exact number of hydrogen atoms to be added to each non-hydrogen atom. 
Once this number is established, the hydrogen atoms can be systematically positioned using theoretical models, taking into account any residual electron density.
Typically, hydrogen atoms are located on the outer edges of a molecule, and their potential to form hydrogen bonds with neighboring molecules serves as an important criterion in determining hydrogen atoms.
As a result, understanding the interaction patterns between molecules becomes essential for fully describing the chemical environment.
We build another TorchMD-Net model for hydrogen atom determination, incorporating joint modeling of both intra-molecular and inter-molecular interactions.
Specifically, we identify equivalent atoms within a $3.2~\mathring{\textrm{A}}$ radius of each non-hydrogen atom, a distance that aligns with the typical maximum bond length of a hydrogen bond, taking into account the crystal's symmetry and periodicity. 
These equivalent atoms serve as auxiliary atoms in the model, introducing necessary intermolecular interaction patterns to predict the number of hydrogen atoms added to each non-hydrogen atom. 
We show that incorporating both intra- and intermolecular interactions yields a significant performance improvement of over $7\%$ compared to using only intramolecular interactions.

\subsection*{CrystalX enables precise, rapid full-atom structure analysis at scale on temporally held-out experimental data}

We curate a large-scale benchmark of 51,334 experimental X-ray diffraction (XRD) patterns from the Crystallography Open Database (COD). The dataset spans organic, organometallic, and inorganic crystals, covering 83 elements across the periodic table and 86 space groups.

To reduce leakage from near-duplicate structures and to better match real deployment, we evaluate CrystalX under a strict time-ordered split protocol. The COD entries span 1994–2024; we train on structures published before 2018 and test on those from 2018--2024, yielding 8,834 held-out samples. This design concentrates the test set on genuinely new structures and mitigates leakage arising from temporal clustering of related compounds. In addition, we perform a random split with an approximate 8:2 train--test ratio; the corresponding results, together with a comparison between the two splitting strategies, are provided in the Supplementary Information.

On the full time-split test set, CrystalX reaches $99.71\%$ accuracy for non-hydrogen atom identification and $99.42\%$ for hydrogen atoms. Performance is consistently strong across all 83 elements, including chemically similar and frequently co-occurring species in organic and organometallic structures, such as carbon (precision: $99.87\%$, recall: $99.91\%$), nitrogen (precision: $99.04\%$, recall: $98.71\%$), oxygen (precision: $99.50\%$, recall: $99.60\%$), and fluorine (precision: $99.72\%$, recall: $99.69\%$). 
Figures~\ref{fig:fig2}a and~\ref{fig:fig2}b show the most common cases, which account for $98.72\%$ of all non-hydrogen atoms and $98.96\%$ of all hydrogen atoms, respectively. Detailed results for both non-hydrogen and hydrogen atom determination are provided in the Supplementary Information.

Because crystallographic models are curated and consumed at the structure level, we further report a structure-integrity metric: the fraction of structures in which all atoms are correctly identified. This metric is deployment-relevant, since a single misassigned atom can break chemical plausibility, distort downstream refinement and analysis, and typically necessitates manual inspection of the entire structure. Under this criterion, CrystalX achieves $94.17\%$ structure-level accuracy for non-hydrogen atoms and $91.79\%$ for hydrogen atoms. The remaining failures are dominated by long-tailed chemical environments---especially for hydrogen---which are sparsely represented in training; we analyze these cases in detail in the Supplementary Information.
Beyond accuracy, CrystalX produces well-calibrated probabilities that enable lightweight, actionable uncertainty handling. We apply a minimal post-processing step: identify the atom with the smallest top-1 vs.\ top-2 probability margin and generate one additional structure-level candidate by flipping this most ambiguous assignment. This single-step correction improves structure-integrity accuracy to $95.80\%$ (non-hydrogen) and $94.35\%$ (hydrogen). Notably, CrystalX processes the entire test set in only a few minutes, substantially exceeding the throughput of manual expert analysis.

\subsection*{Benchmarking CrystalX against existing automated crystallographic tools}
To assess end-to-end practical utility, we integrate CrystalX into a minimal automated pipeline for full-atom structure determination and benchmark against Olex2, a widely used crystallographic software suite. For Olex2, we employ a standard fully automated baseline commonly used in routine workflows: SHELXT for phasing and initial non-hydrogen assignment, SHELXL for refinement, and the Olex2 \texttt{hadd} command for hydrogen placement. All steps run without manual intervention to ensure a fair, automation-to-automation comparison. We also compare against AutoChem (latest ac7) on newly discovered compounds from our routine experimental workflow. AutoChem typically relies on measurement-setup-specific metadata (e.g., instrument and acquisition settings exported from CrysAlisPro~\cite{diffraction2019crysalispro}) that cannot be reconstructed from published CIFs alone; 
By contrast, CrystalX does not rely on such metadata, which constitutes an additional practical advantage.
As a result, a direct AutoChem comparison on the COD test set is not feasible. We evaluate AutoChem under multiple available configurations and report its best-performing results.

On the full COD test set, SHELXT attains $94.81\%$ accuracy for identifying individual non-hydrogen atoms, but its structure-integrity accuracy drops to $46.26\%$. In contrast, CrystalX improves structure integrity by $+47.91$ percentage points. We further stress-test performance under (i) increasing structural complexity (high elemental diversity: C, N, O plus $\geq$5 additional distinct elements) and (ii) decreasing diffraction quality (signal-to-noise ratio, SNR); results are summarized in Fig.~\ref{fig:fig2}cd. As complexity grows, SHELXT degrades sharply and ultimately fails on the largest structures, whereas CrystalX remains substantially more stable. In the most challenging regime ($>$ 90 atoms), where SHELXT collapses, CrystalX still achieves $72.73\%$ structure-level accuracy, and the same probability-guided single-atom correction boosts this to $81.82\%$, highlighting robustness on large, complex crystals. A similar pattern holds under noise: SHELXT is highly SNR-sensitive, while CrystalX is markedly more tolerant, retaining strong performance even at SNR $<$ 5 (85.19\%).

We then evaluate full-atom structure determination in the most discriminative settings---low-SNR measurements and large, compositionally diverse structures---by constructing a challenging benchmark with two subsets: (i) a low-quality subset with SNR $<10$, more than $40$ total atoms, and at least $5$ distinct elements (including C, N, and O; 37 structures); and (ii) a large-structure subset with more than $80$ total atoms and at least $5$ distinct elements (including C, N, and O; 43 structures). Results are summarized in Fig.~\ref{fig:fig2}e. In the low-quality subset, the Olex2 pipeline yields 2 correct structures, whereas CrystalX yields 24. In the large-structure subset, the pipeline produces no correct solutions (0/43), whereas CrystalX solves 22.

To evaluate deployability beyond exact matches, we validate CrystalX predictions with CheckCIF~\cite{platon}, using the published CIFs (expert-solved structures) as ground truth. We focus on stringent, publication-relevant Alert Levels A/B and restrict to Alert Types 2--3, which primarily reflect structural issues and data quality. Among the 24 correctly predicted low-quality structures, 22 exhibit alert severity consistent with the ground truth (i.e., both have no relevant A/B alerts, or both present comparable alerts); among the 22 correctly predicted large structures, 20 match in the same sense. These results support the genuine practical deployability of CrystalX. We also demonstrate that CrystalX can analyze exceptionally large structures---up to 370 non-hydrogen atoms---within seconds (Fig.~\ref{fig:fig2}f,g). Moreover, CrystalX is compatible with Encapsulated Nanodroplet Crystallization~\cite{encapsu-crystal}, an advanced high-throughput crystallization method, underscoring its potential to enable scalable crystallization-to-structure workflows (see Supplementary Information).

Finally, we integrate CrystalX into our day-to-day crystallography pipeline, enabling fully automated, human-free all-atom structure determination for newly discovered compounds (Fig.~\ref{fig:fig2}h; S1, S2). The resulting CIFs pass CheckCIF with no A/B-level alerts, and all key crystallographic indicators fall within expected ranges.
In addition, a human expert independently confirmed the predicted structures through conventional refinement, and the results are consistent with the model’s predictions.
The solved structures have been submitted to the Cambridge Crystallographic Data Centre (CCDC). 
In head-to-head tests on these new compounds, AutoChem (ac7) yields one correct and one incorrect solution, whereas our minimal CrystalX pipeline solves both, with a total runtime of approximately 15~s (vs.\ 20--80+~s for AutoChem). 
Additionally, a broad Internet search identified only one recently published case in which AutoChem could be applied (Fig.~\ref{fig:fig2}h; S3)~\cite{autochem_case}. In that case, AutoChem required $\sim$30~s and still produced an incorrect solution, whereas CrystalX solved sucessfully in less time.
Together, these results demonstrate that CrystalX is practically deployable as a fast, accurate and fully automated solution for crystal structure analysis in real experimental workflows. 
Together, these results—especially under a stringent time-ordered split designed to better reflect prospective real-world deployment—demonstrate that CrystalX is a fast, robust, and fully automated solution for crystal structure analysis in practical experimental workflows.
More broadly, these findings highlight a fundamental advantage of data-driven inference over traditional rule-based pipelines: learned models can capture complex crystallographic patterns more effectively, enabling more robust automated structure determination.

\begin{figure}[!t]
    \centering
    \includegraphics[width=\textwidth, trim={0cm 0cm 0cm 0cm}, clip]{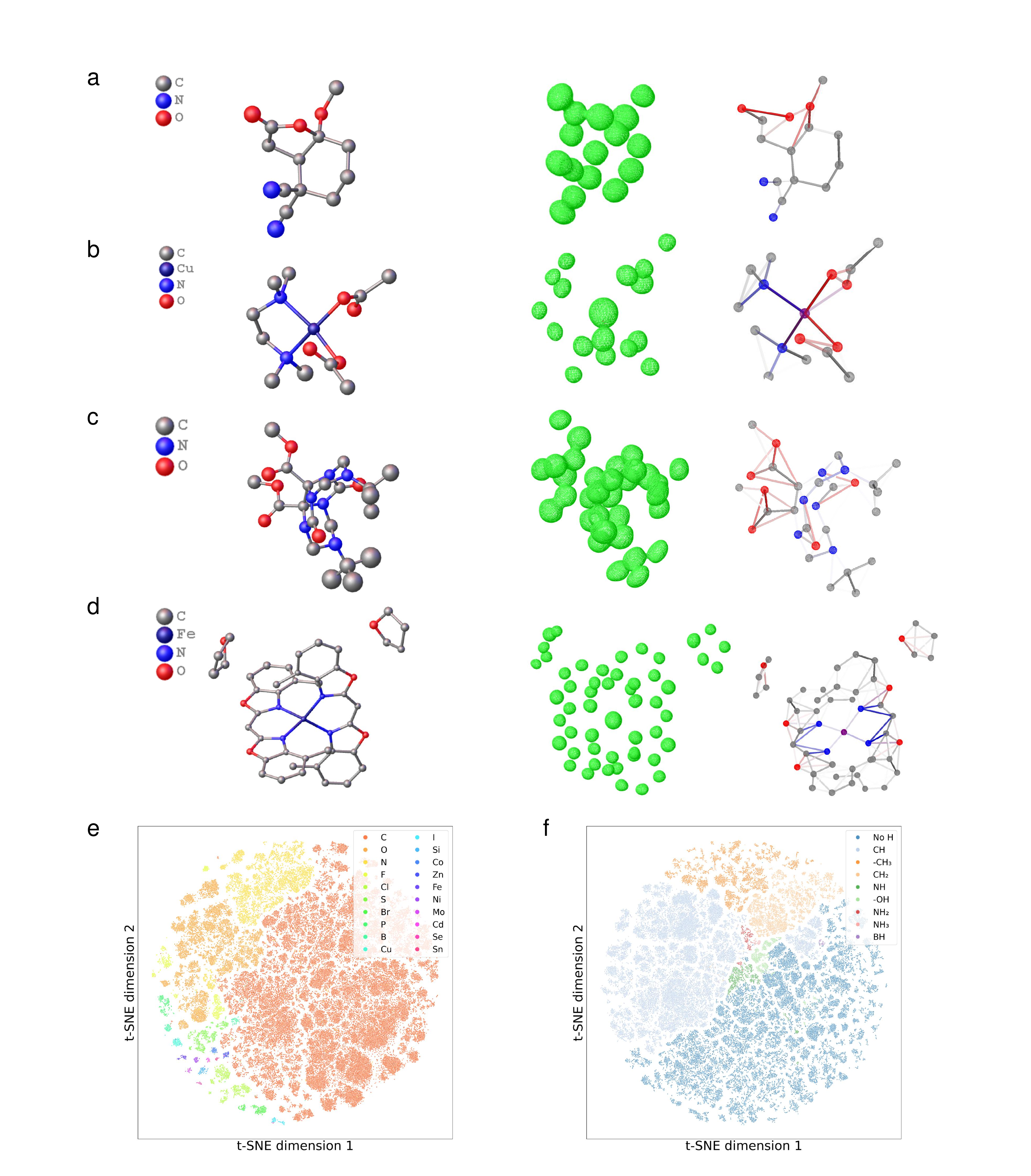}
    \caption{\textbf{Visualization and interpretability of the model's underlying behavior.} (\textbf{a-d}) Visualization of attention maps from Equivalent Transformers using Attention Rollout~\cite{attn-rollout} with class activation demonstrates precise identification of elements like carbon, nitrogen, oxygen, and metals (b,d). This identification is achieved solely by analyzing the geometric patterns of coarse electron density, without the need for any prior information. The interaction intensity is shaded according to class activation levels. (\textbf{e,f}) Two-dimensional t-SNE~\cite{t-sne} projections of the model's learned representations for non-hydrogen atom identification (e) and hydrogen atom identification (f).
    }
    \label{fig:fig3}
\end{figure}

\subsection*{Visualization and interpretability of the model's underlying behavior}
A key question is whether CrystalX merely fits the training distribution, or whether it learns meaningful latent patterns that reflect crystallographic reasoning. To probe the model’s internal behavior, we use Attention Rollout~\cite{attn-rollout} to visualize how the Equivariant Transformer aggregates information during elemental determination from coarse electron density.
Figure~\ref{fig:fig3}a--d illustrates representative organic and organometallic examples. The resulting attention maps highlight the geometric interactions that CrystalX relies on, with color intensity indicating class activation and relative importance. Across diverse chemistries, the model consistently attends to salient local geometry and density features that are critical for distinguishing elements and bonding environments.
We further examine the learned representation space by projecting intermediate embeddings with t-SNE. As shown in Fig.~\ref{fig:fig3}e,f, embeddings cluster into well-separated groups corresponding to different elements and hydrogen types, indicating that CrystalX organizes coarse-density information into a discriminative and interpretable latent geometry that aligns with chemically relevant categories.

\subsection*{CrystalX corrects errors in JCR Q1 journals}

Crystallographic Information Files (CIFs) refined and reported by human experts are generally regarded as the standard for interpreting diffraction data. 
However, could these manual interpretations potentially contain errors?

\begin{figure}[!t]
    \centering
    \includegraphics[width=\textwidth, trim={0cm 5cm 0cm 0cm}, clip]{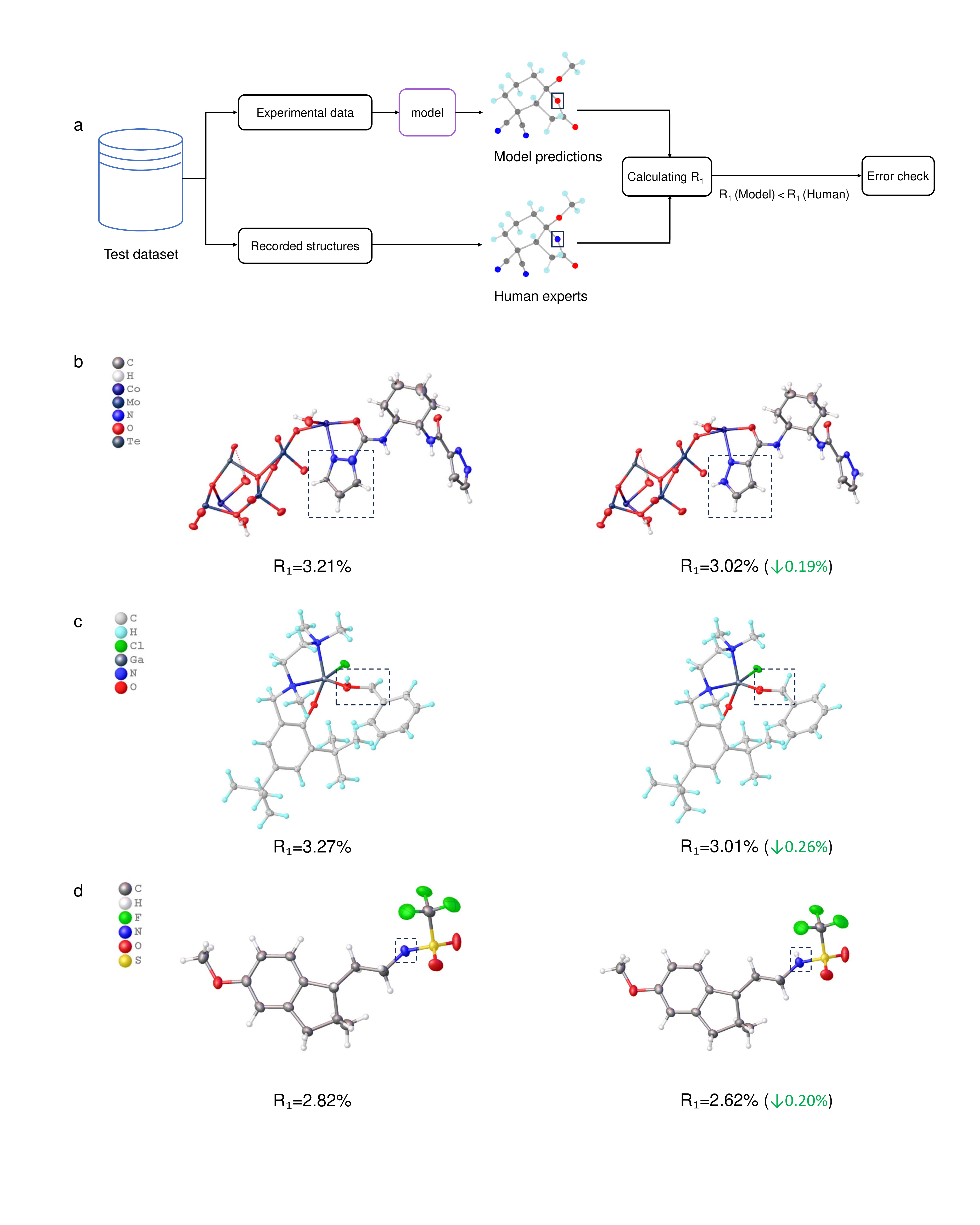}
    \caption{\textbf{CrystalX corrects errors in JCR Q1 journals.} \textbf{a}, Error corrections through crystallographic comparison of the structure analysis results between the model and human experts. \textbf{b-d}, The model corrects three types of errors found in the published literature: (b) misidentification of atoms with similar charges, (c) incorrect placement of hydrogen atoms, and (d) missing hydrogen atoms. These errors are often difficult to detect, passing the rigorous reviews of JCR Q1 journals. The model's analysis offers both improved structural rationality and superior crystallographic metrics.
    }
    \label{fig:fig4}
\end{figure}

When evaluating the model on the test set, we compared crystallographic quality metrics produced by human-reported CIFs and by CrystalX-generated solutions, under identical evaluation conditions (Fig.~\ref{fig:fig4}a). Surprisingly, even among structures published in JCR Q1 journals, the model achieves significantly better crystallographic metrics than the corresponding human solutions.
Manual inspection confirms that these discrepancies arise from expert interpretation errors, whereas the model’s analysis is correct.
Figure~\ref{fig:fig4} highlights three representative cases:
Incorrect identification of non-hydrogen atoms (Fig.~\ref{fig:fig4}b, \texttt{COD\_7244616}), Incorrect hydrogen placement (Fig.~\ref{fig:fig4}c, \texttt{COD\_4345059}), Missing hydrogen atoms (Fig.~\ref{fig:fig4}d, \texttt{COD\_4125905}).
Across the 1,559 JCR Q1 entries in the test set, 10 cases were automatically filtered.
Manual verification confirmed 9 instances of human interpretation errors.

Notably, these errors are hard to detect: in the identified cases \cite{error1,error4,error6,error7,error8}, the corresponding CIFs exhibit no CheckCIF~\cite{platon} A/B-level alerts, despite CheckCIF being widely regarded as a stringent publication standard for crystallographic reporting. We also observe that one error case \cite{error3} matches the outcome produced by Olex2 \texttt{hadd} command, suggesting that the mistake can be systematically induced by commonly used tooling.
Additionally, these incorrect CIFs were published in top-tier (JCR Q1) journals and had already undergone substantial expert review. 
This provides evidence that CrystalX is not only accurate, but also practically valuable—capable of identifying and correcting subtle issues that may evade both routine validation and conventional crystallographic workflows. 
Details are provided in the Supplementary Information.

\section*{Discussions}
The definitive objective of crystal structure analysis is to precisely determine the full atomic structure. 
Recent advances in crystallographic software~\cite{shelxt,shelxs,charge-flipping,altomare1999sir97,sir2014} have largely automated phasing in routine workflows;
the current bottleneck lies in interpreting the resulting coarse electron density.
In practice, crystallographers must manually and iteratively iterate the process of interpreting electron density maps using extensive domain knowledge and software expertise, and the required time and effort scale rapidly with structure size.
This work present the first data-driven approach that uses deep learning to enable high-accuracy routine structure analysis in seconds.
CrystalX is trained and evaluated entirely on real experimental measurements, spanning organic compounds, organometallics, and inorganic materials across 83 elements (nearly the entire periodic table) and 86 space groups.
Unlike prior work that relies heavily on simulated data, this real-data foundation ensures direct applicability to practical crystallographic workflows.
On a strict time-ordered test split of 8,834 real-world COD measurements, CrystalX achieves high accuracy and well-calibrated predictive probabilities, enabling reliable automated structure analysis. Importantly, CrystalX remains robust in the most challenging regimes—very large structures and low-quality (low-SNR) data—where commonly used crystallographic tools degrade sharply, demonstrating genuine capability to generalize to future, previously unseen structures.
CrystalX’s practical value is further underscored by its ability to identify and correct subtle interpretation errors in structures published in JCR Q1 journals—cases that passed expert review and, in some instances, triggered no CheckCIF A/B-level alerts.
Beyond retrospective benchmarks, we demonstrate fully automated structure determination for newly discovered compounds in real experimental settings. In head-to-head comparisons against the latest AutoChem release (ac7), CrystalX successfully solved the structures of two newly discovered compounds and one recently published compound, while also delivering faster performance. By contrast, AutoChem solved only one of the three.
An additional advantage of CrystalX is its ability to bypass the need for precise prior knowledge of each element's content, which human experts typically require and often obtain through additional experimental measurements. 
These strengths make CrystalX ready for practical deployment and position it as a key component of self-driving laboratories~\cite{self-driving-lab}.

X-ray diffraction remains the most widely used technique for routine structure analysis. 
However, recent advancements in novel characterization methods, such as three-dimensional electron diffraction \cite{3ded1,3ded2,3ded3} and serial femtosecond X-ray crystallography \cite{xfel1,3ded-xfel}, have attracted significant attention. 
These techniques are particularly effective for resolving crystal structures with challenging characteristics, including low symmetry, nano- and micro-sized crystals, and radiation sensitivity.
Despite the differences in these characterization techniques, the structure analysis still adheres closely to the standard procedures used in X-ray diffraction~\cite{3ded1,3ded2,3ded3,xfel1,3ded-xfel}.
This implies that CrystalX holds the potential to be extended to these cutting-edge techniques, enabling crystal structure analysis that surpasses traditional X-ray diffraction.

There are long-tail scenarios that occur in crystal structure analysis. One example is rare hydrogen environments associated with uncommon elements: because such configurations appear only sporadically in current datasets, the model encounters too few instances during training, which limits recognition accuracy.
We expect this gap to narrow as training corpora grow and coverage of these rare cases improves.
A substantially harder challenge is crystallographic disorder, which spans a wide variety of heterogeneous, long-tail patterns. Accurately modeling complex disorder is best viewed as a long-horizon, sequential decision-making problem: it requires iterative interpret–refine cycles rather than a single-shot prediction. However, most published structures report only the final refined model. Training a model to handle disorder would therefore require step-level supervision—linking an expert’s intermediate decisions at each disorder site to the corresponding residual electron-density features at specific refinement stages—to create reliable labels. This, in turn, would require access to complete expert refinement trajectories, making data collection and annotation exceptionally difficult. Due to resource constraints, we exclude disordered cases from the present study.
That said, disorder resolution remains a problem of recognizing and reasoning over electron density and residual electron density~\cite{kratzert2015dsr}, which suggests that the geometric deep learning approach used here could be extended to disorder if sufficient annotations were available.
Looking ahead, a promising direction is to leverage recent progress in agentic AI and reinforcement learning for sequential decision-making~\cite{sutton1998reinforcement}.
Disorder modeling naturally fits a sequential decision-making framework, where success depends on executing a series of informed actions—an implicit trajectory—rather than producing a single static output.
In this setting, one could envision an agent-based approach in which an LLM-driven “computer user” interacts with crystallographic tools (e.g., refinement and validation utilities) and relevant references/databases~\cite{achiam2023gpt,yao2022react}. By defining rewards from outcome signals (successful vs. failed disorder resolutions) and collecting trajectories of both successes and failures, reinforcement learning could be used to teach the agent the reasoning patterns required for disorder interpretation. We believe this direction is particularly promising and plan to pursue it in future work.

\section*{Conclusions}
In summary, this work introduce CrystalX, the first deep learning system that fully automates routine crystal structure analysis at full-atom resolution. Trained and evaluated exclusively on large-scale real experimental X-ray diffraction data spanning organic, organometallic, and inorganic crystals across 83 elements and 86 space groups, CrystalX achieves high element-level accuracy and high structure-level integrity under a strict time-ordered split that reflects real deployment on unseen future structures. CrystalX remains robust in the regimes that most often defeat conventional automation---low-SNR measurements and large, compositionally diverse structures---and substantially outperforms the widely used automated Olex2 workflow for end-to-end all-atom structure determination. Its calibrated probabilities further enable lightweight, probability-guided correction that improves structure integrity without expensive search, and its efficiency scales to exceptionally large structures (up to 370 non-hydrogen atoms) within seconds.
Beyond performance, CrystalX provides insight into its underlying behavior: attention visualizations reveal that the model focuses on salient geometric interactions in coarse electron density, and the learned embedding space organizes elements and hydrogen types into well-separated, chemically meaningful clusters. Finally, CrystalX offers practical value beyond routine throughput. By benchmarking against published expert solutions, we show that it can identify and correct subtle interpretation errors even in JCR Q1 journal structures---including cases that pass stringent CheckCIF A/B-level validation and may be systematically induced by commonly used tools. CrystalX is already integrated into our day-to-day crystallography pipeline for newly discovered compounds, enabling fully automated, human-free structure determination and outperforming AutoChem (ac7) in both speed and reliability on our evaluated cases.
These features make CrystalX well-suited to meet the rigorous demands of daily crystal structure analysis and position it for integration into self-driving laboratories, driving forward chemical discovery.

\bibliography{main_ref}
\section*{Supporting Information}
Additional experimental details, materials, and methods (PDF)
\section*{Data and code availability:}
All data required to validate the conclusions of this paper are provided in the manuscript, the supplementary materials, and the GitHub repository (\texttt{https://github.com/kaipengm2/CrystalX}), which includes the relevant data, code, and model parameters.
A web application is available at \texttt{https://crystalx.intern-ai.org.cn/}, enabling users to upload their own data. 
The CrystalX-derived structures for the two newly discovered compounds have been deposited with the Cambridge Crystallographic Data Centre (CCDC) under deposition numbers \texttt{2380405} and \texttt{2380407}.
\section*{Acknowledgments}
This work was done during K.Z.'s internship at Shanghai Artificial Intelligence Laboratory.
We thank Professor Guoyin Yin from Wuhan University for providing two newly discovered compounds for testing.
We also thank Lin Huang and Fanjie Xu for their guidance in using AutoChem.
This work is supported by New Generation Artificial Intelligence-National Science and Technology Major Project (2025ZD0121802) and Intern-Discovery, National Natural Science Foundation of China (No. 62406192), and Shanghai Municipal Special Program for Basic Research on General AI Foundation Models (Grant No. 2025SHZDZX025G03).
\bmhead{Competing interests:}
There are no competing interests to declare.

\end{document}